\documentclass[runningheads]{llncs}
\usepackage{graphicx}
\usepackage{etoolbox,siunitx}
\usepackage{float} 
\usepackage{subfigure}

\usepackage{algorithmic}
\usepackage[linesnumbered,ruled]{algorithm2e}
\usepackage{amsmath}

\let\llncssubparagraph\subparagraph
\let\subparagraph\paragraph
\usepackage[compact]{titlesec}
\let\subparagraph\llncssubparagraph

\begin{document}
\title{A Hybrid Differential Evolution Approach to Designing Deep Convolutional Neural Networks for Image Classification}
%
%
\author{
	Bin Wang \and
	Yanan Sun \and
	Bing Xue \and 
	Mengjie Zhang
}
\titlerunning{A Hybrid DE Approach to Designing CNN for Image Classification}
\authorrunning{B. Wang, Y. Sun, B. Xue and M. Zhang}
%
\institute{School of Engineering and Computer Science\\
	Victoria University of Wellington, PO Box 600, Wellington 6140, NEW ZEALAND
	\email{wangbin@myvuw.ac.nz, \{yanan.sun, bing.xue, mengjie.zhang\}@ecs.vuw.ac.nz}
}
\maketitle              
\begin{abstract}

Convolutional Neural Networks (CNNs) have demonstrated their superiority in image classification, and evolutionary computation (EC) methods have recently been surging to automatically design the architectures of CNNs to save the tedious work of manually designing CNNs. In this paper, a new hybrid differential evolution (DE) algorithm with a newly added crossover operator is proposed to evolve the architectures of CNNs of any lengths, which is named DECNN. There are three new ideas in the proposed DECNN method. Firstly, an existing effective encoding scheme is refined to cater for variable-length CNN architectures; Secondly, the new mutation and crossover operators are developed for variable-length DE to optimise the hyperparameters of CNNs; Finally, the new second crossover is introduced to evolve the depth of the CNN architectures. The proposed algorithm is tested on six widely-used benchmark datasets and the results are compared to 12 state-of-the-art methods, which shows the proposed method is vigorously competitive to the state-of-the-art algorithms. Furthermore, the proposed method is also compared with a method using particle swarm optimisation with a similar encoding strategy named IPPSO, and the proposed DECNN outperforms IPPSO in terms of the accuracy. 

\end{abstract}

\section{Introduction}\label{sec:Introduction}


Convolutional Neural Networks (CNNs) have proved their dominating spot in various machine learning tasks, such as speech recognition \cite{SpeechRecog_Ossama}, sentence classification \cite{Sentence_Yoon} and image classification \cite{ImageNet_Alex}. However, from the existing efforts taken by researchers such as LeNet \cite{LeNetHandWritten_LeCun}\cite{LeNetDocument_LeCun}, AlexNet \cite{ImageNet_Alex}, VGGNet \cite{VGGNet_Simonyan} and GoogLeNet \cite{GoogleNet_Szegedy}, it can be found that designing CNNs for specific tasks could be extremely complicated. 

Since the difficulties of manually designing the architectures of CNNs have been raised more frequently in recent years, 
exploiting evolutionary computation (EC) algorithms to generate deep neural networks \cite{EvolveDNN_yanan} has come into the spotlight to resolve the issues. Interested researchers have accomplished promising results on the automatic design of the architectures of CNNs by using Genetic Programming (GP) \cite{GPNN_Suganuma} and Genetic Algorithms (GAs) \cite{GeneticCNN_Xie}. 
However, the computational cost of the existing algorithms is very expensive, so more research tends to focus on improving the efficiency by developing new algorithms.

Deferential Evolution (DE)  has been proved to be an efficient heuristic for global optimisation over continuous spaces \cite{DE_Storn}, but it has never been used to evolve deep CNNs.
The IP-Based Encoding Strategy (IPES) \cite{ippso_bin} has demonstrated its powerfulness in particle swarm optimisation for evolving deep CNNs, but it has a critical drawback which is that the maximum depth of the CNN architectures has to be set before the commencement of the evolutionary process. Therefore, the encoding strategy is refined in the proposed algorithm to break the constraint of the predefined maximum length. 


\textbf{Goals:}The overall goal of this paper is to explore the ability of DE for automatically evolving the structures and parameters of deep CNNs. The goal will be achieved by designing an effective encoding scheme, new mutation and crossover operators of DE, and a second crossover operator. The proposed method named DECNN will be examined and compared with 12 state-of-the-art methods on six widely-used datasets of varying difficulty. The specific objectives are 

\begin{itemize}
	\item refine the existing effective encoding scheme used by IPPSO \cite{ippso_bin} to break the constraint of predefining the maximum depth of CNNs;
	\item design and develop new mutation and crossover operators for the proposed DECNN method, which can be applied on variable-length vectors to conquer the fixed-length limitation of the traditional DE method;
	\item design and integrate a second crossover operator into the proposed DECNN to produce the children in the next generation representing the architectures of CNNs whose lengths differ from their parents. 
\end{itemize}

\section{Background and Related Work}\label{sec:Background}

\subsection{CNN Architecture}\label{S:background_CNN}

A typical Convolutional Neural Network (CNN) is constituted of four types of layers - convolution layer, pooling layer, fully-connected layer and output layer. The output layer depends only on the specific classification problem. For the example of image classification, the number of classes decides the size of the output layer. Therefore, when designing an architecture of CNNs, the output layer is fixed once the specific task is given. However, to decide the other three types of layers, first of all, the depth of the CNN architecture 
has to be decided; Then, the type of each layer needs to be chosen from convolution layer, pooling layer and fully-connected layer
; Last but no least, since there are different sets of attributes for different types of layers - filter size, stride size and feature maps for the convolution layer; kernel size, stride size and pooling type enclosing max-pooling or average pooling for the pooling layer; and the number of neurons for the fully-connected layer, the attributes of each layer have to be tuned based on its layer type in order to accomplish a CNN architecture that can obtain good performance.


\subsection{Differential Evolution}\label{S:background_DE}

Differential Evolution (DE) is a population-based EC method which searches for the optimal solutions of a problem. 
It has been proved to be a simple and efficient heuristic method for global optimisation over continuous spaces \cite{DE_Storn}. Overall, there are four major steps in a DE algorithm, which are initialisation, mutation, crossover and selection \cite{DE_Algorithm_Kenneth}. First of all, a population of candidate vectors are randomly initialised. Secondly, mutation is applied according to Formula (\ref{eq:DE_mutation}), where $\mathbf{v}_{i,g}$ means the $i\mathit{th}$ temporary candidate vector of the $g\mathit{th}$ generation; $\mathbf{x}_{r0,g}$, $\mathbf{x}_{r1,g}$ and $\mathbf{x}_{r2,g}$ indicate three randomly picked candidates of the $g\mathit{th}$ generation; and $F$ is the differential rate, which is used to control the evolution rate. Thirdly, the crossover is performed based on Formula (\ref{eq:DE_crossover}), where $u_{j,i,g}$ represents the $j\mathit{th}$ dimension of the $i\mathit{th}$ candidate at the $g\mathit{th}$ generation. At the beginning of the crossover process for each candidate, a random number $j_{rand}$ is generated, and then for each dimension of each candidate vector, another random number $rand_{j}$ is generated, which then is compared with the crossover rate $Cr$ and $j_{rand}$ as shown in Formula (\ref{eq:DE_crossover}) to decide whether the crossover applies on this dimension. After applying the DE operators, a trial vector $\mathbf{u}_{i,g}$ is produced, which is then compared with the parent vector 
to select the one that has a better fitness. By iterating the steps of mutation, crossover and selection until the stopping criterion is met, the best candidate can be found. 

\begin{equation}\label{eq:DE_mutation}
\mathbf{v}_{i,g} = \mathbf{x}_{r0,g} + F \times (\mathbf{x}_{r1,g}-\mathbf{x}_{r2,g})
\end{equation}
\vspace{-3mm}
\begin{equation}\label{eq:DE_crossover}
u_{j,i,g} = 
\begin{cases}
v_{j,i,g} & \text{if $rand_{j}(0,1) < Cr$ or $j=j_{rand}$} \\
x_{j,i,g} & \text{otherwise}
\end{cases}
\end{equation}

\subsection{Related Work}

Recently, more and more research has been done using EC methods to evolve the architectures of CNNs. Genetic CNN \cite{GeneticCNN_Xie} and CGP-CNN \cite{GPNN_Suganuma} are two of the most recent proposed methods that have achieved promising results in comparison with the state-of-the-art human-designed CNN architectures. 

Genetic CNN uses a fixed-length binary string to encode the connections of CNN architectures in a constrained case. It splits a CNN architecture into stages. Each stage is comprised of numerous convolutional layers which may or may not connect to each other, and pooling layers are used between stages to connect them to construct the CNN architecture.  Due to the fixed-length binary representation, the number of stages and the number of nodes in each stage have to be predefined, so a large fraction of network structures are not explored by this algorithm. Other than that, the encoding scheme of Genetic CNN only encodes the connections, i.e. whether two convolutional layers are connected or not; while the hyperparameters of the convolutional layers, e.g. the kernel size, the number of feature maps, and the stride size, are not encoded, so Genetic CNN does not have the ability to optimise the hyperparameters.

CGP-CNN utilises Cartesian Genetic Programming (CGP) \cite{CGP_Miller} because the flexibility of CGP's encoding scheme is suitable to effectively encode the complex CNN architectures. CGP-CNN employs a matrix of $N_{r}$ rows and $N_{c}$ columns to represent the layers of a CNN architecture and their connections, respectively, so the maximum number of layers is predefined. In addition, as six types of node functions called ConvBlock, ResBlock, max pooling, average pooling, concatenation and summation are prepared, CGP-CNN is confined to explore the limited types of layers of CNN architectures. Last but not least, from the experimental results, the computational cost of CGP-CNN is quite high because training CNNs in fitness evaluation is time-consuming. 

In summary, manually design of CNN architectures and parameters is very challenging and time-consuming. Automatically evolving the architectures of deep CNNs is a promising approach, but their potential has not been fully explored. DE has shown as an efficient method in global optimisation, but has not been used to evolve deep CNNs. Therefore, we would like to investigate a new approach using DE to automatically evolve the architectures of deep CNNs.

\section{The Proposed Algorithm}\label{sec:ProposedAlgorithm}


The proposed DECNN method uses DE as the main evolutionary algorithm, 
and a second crossover operator is proposed to generate children whose lengths differ from their parents to fulfil the requirement of evolving variable-length architectures of CNNs. 


\subsection{DECNN Algorithm Overview}\label{S:algorithms_DECNN_overview}

The overall procedure of the proposed DECNN algorithm 
is written in Algorithm \ref{alg:iphdg_framework}.

\vspace{-6mm}
\begin{algorithm}[ht]
	\caption{Framework of IPDE}
	\label{alg:iphdg_framework}
	\begin{algorithmic}
		\renewcommand{\algorithmicrequire}{\textbf{Input:}}
		\renewcommand{\algorithmicensure}{\textbf{Output:}}
		\STATE $P \leftarrow$ Initialise the population elaborated in Section \ref{S:algorithms_population_init};
		\STATE $P\_best \leftarrow empty$;
		\WHILE{termination criterion is not satisfied}
		\STATE Apply the refined DE mutation and crossover described in Section \ref{S:algorithms_DECNN_update};
		\STATE Apply the proposed second crossover to produce two children, and select the best between the two children and their parents illustrated in Section \ref{S:algorithms_DECNN_gacrossover};
		\STATE evaluate the fitness value of each individual;
		\STATE $P\_best \leftarrow$ retrieve the best individual in the population;
		\ENDWHILE
	\end{algorithmic}
\end{algorithm}
\vspace{-6mm}

\subsection{Adjusted IP-Based Encoding Strategy}\label{S:algorithms_IPencoding}


The proposed IP-Based Encoding Strategy is to use one IP Address to represent one layer of Deep Neural Networks (DNNs) and push the IP address into a sequence of interfaces, each of which bears an IP address and its corresponding subnet, in the same order as the order of the layers in DNNs. Taking CNNs as an example, the typical CNNs are composed of three types of layers - convolutional layer, pooling layer and fully-connected  layer. 
The first step of the encoding is to work out the range that can represent each attribute of each type of the CNN layer. There are no specific limits for the attributes of CNN layers, but in order to practically apply optimisation algorithms on the task, each attribute has to be given a range which has enough capacity to achieve an optimal accuracy on the classification problems. In this paper, the constraints for each attribute are designed to be capable of accomplishing a relatively low error rate. To be specific, for the convolutional layer, there are three attributes, which are filter size ranging from 1 to 8, number of feature maps from 1 up to 128, and the stride size with the range from 1 to 4. As the three attributes need to be combined into one number, a binary string with 12 bits can contain all the three attributes of the convolutional layer, which are 3 bits for filter size, 7 bits for the number of feature maps, and 2 bits for the stride size. Following the similar way, the pooling layer and fully-connected  layer can be carried in the binary strings with 5 bits and 11 bits, respectively. The details of the range of each attribute are listed in Table \ref{table:CNNFields}. 

\begin{table}[ht]
	\renewcommand{\arraystretch}{1.3}
	\tiny
	\caption{The ranges of the attributes of CNN layers - Convolutional, Pooling, Fullly-connected layer}
	\vspace{-4mm}
	\label{table:CNNFields}
	\centering
	\begin{tabular}{|c|c|c|c|}
		\hline
		Layer Type & Parameter & Range & \# of Bits\\
		\hline
		Conv & Filter size & [1,8] & 3\\
		\hline
		& \# of feature maps & [1,128] & 7\\
		\hline
		& Stride size & [1,4] & 2\\
		\hline
		& \textbf{Total} &  & 12\\
		\hline
		\hline
		Pooling & Kernel size & [1,4] & 2\\
		\hline
		& Stride size & [1,4] & 2\\
		\hline
		& Type: 1(maximal), 2(average) & [1,2] & 1\\
		\hline
		& \textbf{Total} &  & 5\\
		\hline
		\hline
		Fully-connected & \# of Neurons & [1,2048] & 11\\
		\hline
		& \textbf{Total} &  & 11\\
		\hline
	\end{tabular}
\end{table}

\begin{figure}[ht]
	\centering
	\includegraphics[width=.5\linewidth]{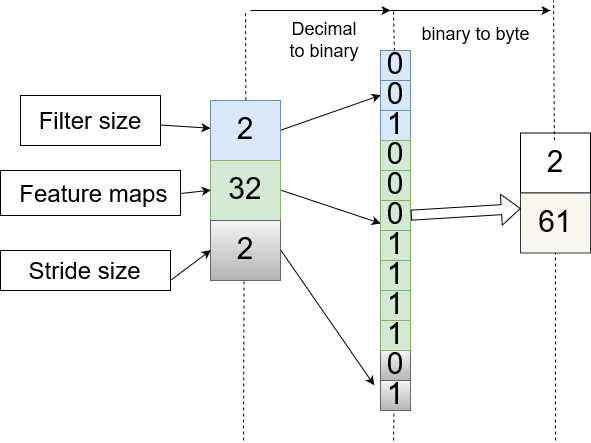}
	\vspace{-4mm}
	\caption{An example of how to encode a convolutional layer using a byte array}
	\label{fig:ip_encoding_conv}
\end{figure}

Once the number of bits of the binary strings has been defined, a specific CNN layer can be easily translated to a binary string. Suppose a convolutional layer with the filter size of 2, the number of feature maps of 32 and the stride size of 2 is given, the corresponding binary strings of [001], [000 1111] and [01] can be calculated by converting the decimal numbers\footnote{Before the conversion, 1 is subtracted from the decimal number because the binary string starts from 0, while the decimal value of the attributes of CNN layers begins with 1} to the corresponding binary numbers. The final binary string that stands for the given convolutional layer is [001 000 1110 01] by joining the binary strings of the three attributes together. The details of the example are shown in Fig. \ref{fig:ip_encoding_conv}. 

\begin{figure}[ht]
	\centering
	\includegraphics[width=.5\linewidth]{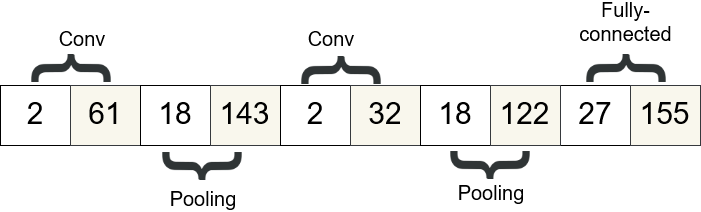}
	\vspace{-3mm}
	\caption{An example of the encoded vector of a CNN architecture}
	\label{fig:ip_encoding_full_vec}
\end{figure}

Similar like network engineering where the subnet has to be defined before allocating an IP address to an interface, i.e. a laptop or desktop, the IP-Based Encoding Strategy needs to design a subnet for each type of CNN layers. 
Since the number of bits of each layer type decides its size of the search space, and the pooling layer takes much fewer bits than the other two, the chances of a pooling layer being chosen would be much smaller than the other two.
In order to balance the probability of each layer type being selected, a place-holder of 6 bits is added to the binary string of the pooling layer to make it 11 bits, which brings the odds of picking a pooling layer the same as that of a fully-connected  layer. As there are three types of layers with the maximum bits of 12, a 2-byte binary string has sufficient capacity to bear the encoded CNN layers. Starting with the convolutional layer of 12 bits, as this is the first subnet, the 2-byte binary representation of the starting IP address would be [0000 0000 0000 0000], and the finishing IP address would be [0000 1111 1111 1111]; The fully-connected  layer of 11 bits starts from the binary string [0001 0000 0000 0000] by adding one to the last IP address of the convolutional layer, and ends to [0001 0111 1111 1111]; And similarly, the IP range of the pooling layer can be derived - from [0001 1000 0000 0000] to [0001 1111 1111 1111]. The IP ranges of the 2-byte style for each subset are shown in Table \ref{table:Subnets}, which are obtained by converting the aforementioned binary strings to the 2-byte strings. Now it is ready to encode a CNN layer into an IP address, and the convolutional layer detailed in Fig. \ref{fig:ip_encoding_conv} is taken as an example. The binary representation of the IP address is [0000 0010 0011 1001] by summing up the binary string of the convolutional layer and the starting IP address of the convolutional layer's subnet, which can be converted to a 2-byte IP address of [2.61]. Fig. \ref{fig:ip_encoding_full_vec} shows an example vector encoded from a CNN architecture with 2 convolutional layers, 2 pooling layers and 1 fully-connected layer. 

\vspace{-6mm}

\begin{table}[ht]
	\renewcommand{\arraystretch}{1.3}
	\tiny
	\caption{Subnets distributed to the three types of CNN layers and the disabled layer}
	\vspace{-3mm}
	\label{table:Subnets}
	\centering
	\begin{tabular}{|c|c|}
		\hline
		Layer type & IP Range\\
		\hline
		Convolutional Layer & 0.0-15.255\\
		\hline
		fully-connected  layer & 16.0-23.255\\
		\hline
		pooling layer & 24.0-31.255\\
		\hline
	\end{tabular}
\end{table}

\vspace{-6mm}
\subsection{Population Initialisation}\label{S:algorithms_population_init}

As the individuals are required to be in different lengths, the population initialisation starts by randomly generating the lengths of individuals. In the proposed DECNN, the length is randomly sampled from a Gaussian distribution with a standard deviation $\rho$ of 1 and a centre $\mu$ of a predefined length depending on the complexity of the classification task as shown in Equation (\ref{eq:gaussian_distribution}). After obtaining the candidate's length, the layer type and the attribute values can be randomly generated for each layer in the candidate. By repeating the process until reaching the population size to accomplish the population initialisation. 

\vspace{-2mm}
\begin{equation}\label{eq:gaussian_distribution}
P(x) = \frac{1}{{\sigma \sqrt {2\pi } }}e^{{{ - \left( {x - \mu } \right)^2 } \mathord{\left/ {\vphantom {{ - \left( {x - \mu } \right)^2 } {2\sigma ^2 }}} \right. \kern-\nulldelimiterspace} {2\sigma ^2 }}}
\end{equation}
\vspace{-2mm}

\subsection{Fitness Evaluation}\label{S:algorithms_fitness_evaluation}


The fitness evaluation process is illustrated in Algorithm \ref{alg:fitness}. First of all, four arguments are taken in by the fitness evaluation function, which are the candidate solution which represents an encoded CNN architecture, the training epoch number for training the model decoded from the candidate solution, the training set which is used to train the decoded CNN architecture, and the fitness evaluation dataset on which the trained model is tested to obtain the accuracy used as the fitness value. Secondly, the fitness evaluation process is pretty straightforward by using the back propagation to train the decoded CNN architecture on the training set for a fixed number of epochs, and then obtaining the accuracy on the fitness evaluation set, which is actually used as the fitness value. For the purpose of reducing computational cost, the candidate CNN is only trained on a partial dataset for a limited number of epochs, which are controlled by the arguments of the fitness function - $k$, $D\_train$ and $D\_fitness$. 

\vspace{-2mm}
\begin{algorithm}[ht]
	\caption{Fitness Evaluation}
	\label{alg:fitness}
	\begin{algorithmic}
		\renewcommand{\algorithmicrequire}{\textbf{Input:}}
		\renewcommand{\algorithmicensure}{\textbf{Output:}}
		\REQUIRE The candidate solution $c$, the training epoch number $k$, the training set $D\_train$, the fitness evaluation dataset $D\_fitness$;
		\ENSURE The fitness value $fitness$;
		\STATE Train the connection weights of the CNN represented by the candidate $c$ on the training set $D\_train$ for $k$ epochs;
		\STATE $acc \leftarrow $ Evaluate the trained model on the fitness evaluation dataset $D\_fitness$
		\STATE $fitness \leftarrow acc$;
		\RETURN $fitness$
	\end{algorithmic}
\end{algorithm}

\subsection{DECNN DE Mutation and Crossover}\label{S:algorithms_DECNN_update}

The proposed DECNN operations are similar to the standard DE mutation and crossover as described in Section \ref{S:background_DE}, but it introduces an extra step to trim the longer vectors before applying any operation because the DECNN candidates have different lengths and the traditional DE operations in Equation (\ref{eq:DE_mutation}) and (\ref{eq:DE_crossover}) only apply on fixed-length vectors. To be specific, the three random vectors for the mutation are trimmed to the shortest length of them, and during the crossover, if the trial vector generated by the mutation is longer than the parent, it will be trimmed to the length of the parent.

\subsection{DECNN second crossover}\label{S:algorithms_DECNN_gacrossover}

\begin{figure}[ht]
	\centering
	\includegraphics[width=.7\linewidth]{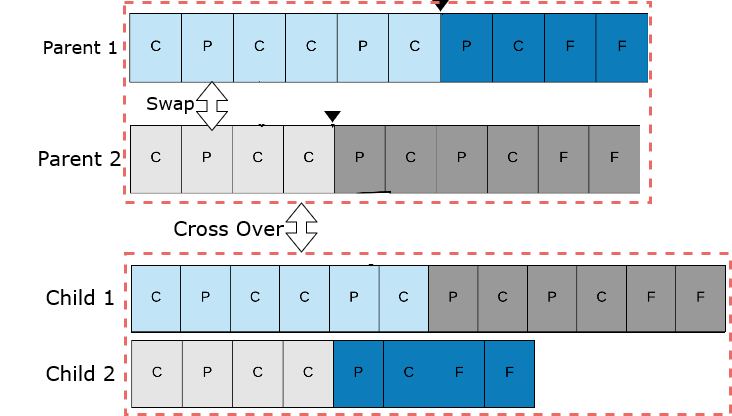}
	\vspace{-3mm}
	\caption{second crossover of the proposed DECNN algorithm}
	\vspace{-6mm}
	\label{fig:GA_liked_crossover}
\end{figure}

Similar as the crossover of GAs, each individual of the two parents is split into two parts by slicing the vector at the cutting points, and swap one part with each other. The cutting point is chosen by randomly finding a position based on Gaussian distribution with the middle point as the centre and a hyperparameter $\rho$ as the standard deviation to control the variety in the population. The flow of the second crossover is outlined in Fig. \ref{fig:GA_liked_crossover}.

\vspace{-6mm}

\section{Experiment design}\label{sec:EPDesign}

\subsection{Benchmark Datasets}

In the experiments, six widely-used benchmark datasets\footnote{Download URL:http://www.iro.umontreal.ca/~lisa/twiki/bin/view.cgi/Public/MnistVariations} \cite{DLBaseline_Chan} are chosen to examine the proposed algorithm, which are the datasets of MNIST Basic (MB), MNIST with a black and white image as the Background Image (MBI),  MNIST Digits Rotated with a black and white image as the Background Image (MDRBI), MNIST with a Random Background (MRB), MNIST with Rotated Digits (MRD), and CONVEX. The MB benchmark dataset and its four variants, the MBI, MDRBI, MRB and MRD datasets, consist of handwritten digits and the corresponding labels from 0 to 9, and each of the datasets is composed of a training set of 12,000 instances and a test set of 5,000 instances; while convex images and non-convex images with the corresponding labels constitute the CONVEX dataset, which is split into a training set of 8,000 examples and a test set of 5,000 examples. Each image in these benchmark datasets has $28 \times 28$ pixels. 
The reason for picking the six aforementioned datasets is to fulfil the purpose of thoroughly testing the proposed algorithms because both multi-class classification tasks for MB and its variants and the binary classification tasks for CONVEX are included in the selected datasets, and the complexity of MB and its variants differ from each other where MB is the simplest one; while MDRBI is the most complicated.

\subsection{State-of-the-art Competitors}

Six state-of-the-art methods are reported to have achieved promising results on the aforementioned benchmark datasets in the literature \cite{DLBaseline_Chan}. Therefore, they are picked as the peer competitors of the proposed algorithm, which are CAE-2 \cite{CAE_Rifai}, TIRBM \cite{TIRBM_Sohn}, PGBM+DN1 \cite{PGBMDN1_Sohn}, ScatNet-2 \cite{ScatteringCNN_Bruna}, RandNet-2 \cite{DLBaseline_Chan}, PCANet-2 (softmax) \cite{DLBaseline_Chan}, LDANet-2 \cite{DLBaseline_Chan}, SVM+RBF \cite{DeepArchitectureEval_Larochelle}, SVM+Poly \cite{DeepArchitectureEval_Larochelle}, NNet \cite{DeepArchitectureEval_Larochelle}, SAA-3 \cite{DeepArchitectureEval_Larochelle} and DBN-3 \cite{DeepArchitectureEval_Larochelle}.

\subsection{Parameter settings of the proposed EC methods}\label{S:ec_parameters}

All of the parameters are configured according to the conventions in the communities of DE \cite{DEParameter_Gamperle} along with taking into account a small population to safe computation time and the complexity of the search space. For the evolutionary process, 30 is set as the population size and 20 is used as the number of generations; In regard to the fitness evaluation, the number of training epochs is set to 5 and 10\% of the training dataset is passed for evaluation; In terms of the DE parameters, 0.6 and 0.45 are used as the differential rate and the crossover rate, respectively; The hyperparameter $\rho$ of second crossover is set to 2, and $\mu$ of the population initialisation is set to 10; 30 independent runs is performed by the proposed DECNN on each of the benchmark dataset. 

\section{Results and analysis}\label{sec:EPResults}


Since DE is stochastic, statistical significance test is required to make the comparison result more convincing. When comparing the proposed DECNN with the state-of-the-art methods, One Sample T-Test is applied to test whether the results of DECNN is better; when the comparison of error rates between DECNN and the peer EC competitor named IPPSO \cite{ippso_bin} is performed, Two Sample T-test is utilised to determine whether the difference is statistically significant or not. 
Table \ref{table:DECNN_VS_SOA} shows the comparison results between the proposed DECNN and the state-of-the-art algorithms; Table \ref{table:IPDE_VS_IPPSO} compares DECNN with IPPSO.

\subsection{DECNN vs. State-of-the-Art methods}

The experimental results and the comparison between the proposed DECNN and the state-of-the-art methods are shown in Table \ref{table:DECNN_VS_SOA}. In order to clearly show the comparison results, the terms (+) and (-) are provided to indicate the result of DECNN is better or worse than the best result obtained by the corresponding peer competitor; The term (=) shows that the mean error rate of DECNN are slightly better or worse than the competitor, but the difference is not significant from the statistical point of view; The term -- means there are no available results reported from the provider or cannot be counted. 

It can be observed that the proposed DECNN method achieves encouraging performance in terms of the error rates shown in Table \ref{table:DECNN_VS_SOA}. To be specific, the proposed DECNN ranks the fifth on both the CONVEX and MB benchmark datasets; for the MBI benchmark, DECNN beats all of the state-of-the-art methods; for the MDRBI dataset, the mean error rate of DECNN is the fourth best, but the P-value of One Sample T-Test between DECNN and the third best is 0.0871, which indicates that the significance difference is not supported from the statistical point of view, so DECNN ties the third with PGBM+DN-1; for the MRB benchmark, the mean error rate of DECNN is smaller than all other methods, but the difference between DECNN and the second best algorithm is not significant given the calculated P-value of 0.1053, so DECNN ties the first with PGBM+DN-1; for the MRD benchmark, DECNN outruns the state-of-the-arts method apart from TIRBM. In addition, by looking at the best results of DECNN, DECNN achieves the smallest error rates on five out of the six datasets compared with the 12 state-of-the-art methods, which are \num{1.03200287818908}\% on MB, \num{5.66599760055542}\% on MBI, \num{32.85199904}\% on MDRBI, \num{3.458004236}\% on MRB and \num{4.066001415}\% on MRD. This shows that DECNN has the potential to improve the state-of-the-art results. 

\begin{table}[ht]
	\renewcommand{\arraystretch}{1.3}
	\tiny
	\vspace{-6mm}
	\caption{The classification errors of DECNN against the peer competitors}
	\vspace{-3mm}
	\label{table:DECNN_VS_SOA}
	\centering
	\begin{tabular}{|m{2cm}|S[table-format=2.2] c|S[table-format=2.2] c|S[table-format=2.2] c|S[table-format=2.2] c|S[table-format=2.2] c|S[table-format=2.2] c|}
		\hline 
		classiﬁer & & CONVEX & & MB & & MBI & & MDRBI & & MRB & & MRD \\ 
		\hline 
		CAE-2  & & -- & 2.48 & (+) & 15.5 & (+) & 45.23 & (+) & 10.9 & (+) & 9.66 & (+) \\
		\hline 
		TIRBM  & & -- &  & -- &  & --  & 35.5 & (-) &  & -- & 4.2 & (-)\\
		\hline 
		PGBM+DN-1  & & -- &  & -- & 12.15 & (+) & 36.76 & (=) & 6.08 & (=) & & --\\
		\hline 
		ScatNet-2  & 6.5 & (-) & 1.27 & (-) & 18.4 & (+) & 50.48 & (+) & 12.3 & (+) & 7.48 & (+)\\
		\hline 
		RandNet-2  & 5.45 & (-) & 1.25 & (-) & 11.65 & (+) & 43.69 & (+) & 13.47 & (+) & 8.47 & (+)\\
		\hline 
		PCANet-2 (softmax)  & 4.19 & (-) & 1.4 & (-) & 11.55 & (+) & 35.86 & (-) & 6.85 & (+) & 8.52 & (+)\\
		\hline 
		LDANet-2  & 7.22 & (-) & 1.05 & (-) & 12.42 & (+) & 38.54 & (+) & 6.81 & (+) & 7.52 & (+)\\
		\hline 
		SVM+RBF  & 19.13 & (+) & 30.03 & (+) & 22.61 & (+) & 55.18 & (+) & 14.58 & (+) & 11.11 & (+)\\
		\hline 
		SVM+Poly  & 19.82 & (+) & 3.69 & (+) & 24.01 & (+) & 54.41 & (+) & 16.62 & (+) & 15.42 & (+)\\
		\hline 
		NNet & 32.25 & (+) & 4.69 & (+) & 27.41 & (+) & 62.16 & (+) & 20.04 & (+) & 18.11 & (+)\\
		\hline 
		SAA-3 & 18.41 & (+) & 3.46 & (+) & 23 & (+) & 51.93 & (+) & 11.28 & (+) & 10.3 & (+)\\ 
		\hline 
		DBN-3 & 18.63 & (+) & 3.11 & (+) & 16.31 & (+) & 47.39 & (+) & 6.73 & (+) & 10.3 & (+)\\ 
		\hline 
		\hline 
		DECNN(best) & 7.99200534820556 &  & 1.03200287818908 & & 5.66599760055542 & & 32.85199904 & & 3.458004236 & & 4.066001415 & \\ 
		\hline 
		DECNN(mean) & 11.1921056069825 &  & 1.45690014362335 & & 8.68529417 & & 37.55115771 & & 5.557766485 & & 5.53253363 & \\ 
		\hline 
		DECNN(standard deviation) & 1.94301999129248 & & 0.113365088 & & 1.405402073 & & 2.446911498 & & 1.711009538 & & 0.4472713045 & \\ 
		\hline 
	\end{tabular}
\end{table}

\vspace{-6mm}
\subsection{DECNN vs. IPPSO}

In Table \ref{table:IPDE_VS_IPPSO}, it can be observed that by comparing the results between DECNN and IPPSO, the mean error rates of DECNN are smaller across all of the six benchmark datasets, and the standard deviations of DECNN is less than those of IPPSO on five datasets out of the six, so the overall performance of DECNN is superior to IPPSO. 
The second crossover operator improves the performance of DECNN because it performs a kind of local search between the two children and their parents both in terms of the depth of CNN architectures and their parameters. 

\begin{table}[ht]
	\renewcommand{\arraystretch}{1.3}
	\tiny
	\vspace{-6mm}
	\caption{Classification rates of DECNN and IPPSO}
	\vspace{-3mm}
	\label{table:IPDE_VS_IPPSO}
	\centering
	\begin{tabular}{|m{2cm}|S[table-format=2.2] c|S[table-format=2.2] c|S[table-format=2.2] c|S[table-format=2.2] c|S[table-format=2.2] c|S[table-format=2.2] c|}
		\hline
		 & & CONVEX & & MB & & MBI & & MDRBI & & MRB & & MRD \\ 
		\hline
		DECNN(mean) & 11.1921056069825 &  & 1.45690014362335 & & 8.68529417 & & 37.55115771 & & 5.557766485 & & 5.53253363 & \\ 
		\hline 
		DECNN(standard deviation) & 1.94301999129248 & & 0.113365088 & & 1.405402073 & & 2.446911498 & & 1.711009538 & & 0.4472713045 & \\ 
		\hline 
		IPPSO(mean) & 12.64506698 &  & 1.558134754 & & 9.857467214 & & 38.79073193 & & 6.255199909 & & 6.071532965 & \\ 
		\hline 
		IPPSO(standard deviation) & 2.13484989 & & 0.1704976496 & & 1.835422058 & & 5.379980526 & & 1.542705767 & & 0.7121172245 & \\ 
		\hline 
		P-value & \textbf{0.01} & & \textbf{0.01} & & \textbf{0.01} & & 0.2554 & & 0.1027 & & \textbf{0.001} & \\ 
		\hline
	\end{tabular}
\end{table}

\vspace{-6mm}
\subsection{Evolved CNN Architecture}

After examining the evolved CNN architectures, it is found that DECNN demonstrates its capability of evolving the length of the architectures. When the evolutionary process starts, the lengths of individuals are around 10; while the lengths of evolved CNN architectures drop to 3 to 5 depending on the complexity of the datasets, which proves that DECNN has the ability of effectively evolving CNN architectures of any lengths.

\section{Conclusions and Future Work}\label{sec:Conclusion}

The goal of this paper is to develop a novel DE-based algorithm to automatically evolve the architecture of CNNs for image classification without any constraint of the depth of CNN architectures. This has been accomplished by designing and developing the proposed hybrid differential evolution method. More specifically, three major contributions are made by the proposed DECNN algorithm. First of all, the IP-Based Encoding Strategy has been improved by removing the maximum length of the encoded vector and the unnecessary disabled layer in order to achieve a real variable-length vector of any length; Secondly, the new DE operations - mutation, crossover are developed, which can be applied to candidate vectors of variable lengths; Last but not least, a novel second crossover is designed and added to DE to produce children having different lengths from their parents. The second crossover plays an important role to search the optimal depth of the CNN architectures because the two children created through the second crossover have different length from their parents - one is longer and the other is shorter, and during the selection from the two children and the two parents, the candidate with a better fitness survives to the next generation, which indicates that the length of the remaining candidate tends to be better than the other three. 

The proposed DECNN has achieved encouraging performance. By comparing the performance of DECNN with the 12 state-of-the-art competitors on the six benchmark datasets, it can be observed that DECNN obtains a very competitive accuracy by ranking the first on the MBI and MRB datasets, the second and the third on the MRD and MDRBI datasets, respectively, and the fifth on the MB and CONVEX datasets. 
In a further comparison with the peer EC competitor, the best results are achieved by DECNN on five out of the six datasets.

There are a couple of potential future works that can be done based on the proposed DECNN. As can been seen, the DECNN method gains much better accuracy on the most complex benchmark among all of the six benchmark datasets, which implies DECNN is very likely to perform well on large and complex datasets, so it is worthy investigating the DECNN algorithm on larger and more complex datasets in order to obtain an insight of how it will perform for industrial usage. 

%
%
%
\def\url#1{}
\def\doi#1{}
\bibliographystyle{splncs04}
\bibliography{IPDE}
\end{document}